\documentclass[10pt,twocolumn,letterpaper]{article}
\usepackage[accsupp]{axessibility}
\usepackage{iccv}
\usepackage{times}
\usepackage{epsfig}
\usepackage{amsmath,amsthm,amssymb,amsfonts,bbm,stmaryrd}
\usepackage{enumitem}
\usepackage{lipsum}
\usepackage{float}
\usepackage{textcomp}
\usepackage{multirow} %
\usepackage{graphicx} %
\usepackage{subcaption}
\usepackage{booktabs} %
\usepackage[normalem]{ulem}
\usepackage{makecell}
\usepackage{courier}
\usepackage{bbm}
\usepackage{algorithm,algpseudocode}
\usepackage{adjustbox}
\usepackage{tabularx}
\usepackage{booktabs}
\usepackage{setspace}

\usepackage{threeparttable}
\usepackage{makecell}
\usepackage{cancel}
\usepackage{times}
\usepackage{latexsym}
\usepackage{graphicx}
\usepackage{xspace}
\usepackage{dirtytalk}
\usepackage{amsmath}
\usepackage{csquotes}
\usepackage{graphicx}
\usepackage{multirow}
\usepackage{makecell}
\usepackage{booktabs}
\usepackage{pgfplots}
\usepackage{enumitem}
\usepackage{amsfonts}
\usepackage{tabularx}
\usepackage{longtable}
\usepackage{subcaption}
\usepackage{pifont}
\usepackage[square,numbers]{natbib}
\newcommand{\cmark}{\ding{51}}
\newcommand{\xmark}{\ding{55}}
\newcommand{\modelEmoji}{\includegraphics[height=.9em,trim=0 .4em 2em 0]{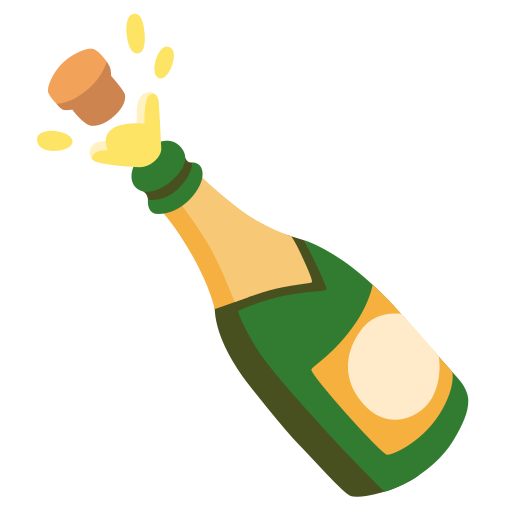}}
\newcommand{\dataEmoji}{\includegraphics[height=.9em,trim=0 5em 2em 0]{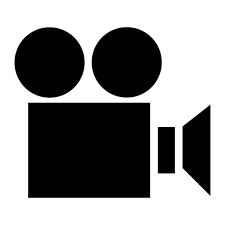}}
\newcommand{\champagne}{\modelEmoji\xspace}
\newcommand{\youtube}{\dataEmoji\xspace}

\newcommand{\datasetName}{\youtube\textsc{YTD-18M}\xspace}
\newcommand{\modelName}{\champagne\textsc{Champagne}\xspace}
\newcommand{\datasetNameNoEmoji}{\textsc{YTD-18M}\xspace}
\newcommand{\modelNameNoEmoji}{\textsc{Champagne}\xspace}
\newcommand{\modelNameNoEmojiBase}{\textsc{Champagne-Base}\xspace}
\newcommand{\modelNameNoEmojiLarge}{\textsc{Champagne-Large}\xspace}
\newcommand{\modelNameNoEmojiXL}{\textsc{Champagne-XL}\xspace}

\usepackage[breaklinks=true,bookmarks=false]{hyperref}

\iccvfinalcopy %

\def\iccvPaperID{5772} %

\ificcvfinal\fi

\begin{document}

\title{\modelName: Learning Real-world Conversation \\ from Large-Scale Web Videos}

\author{
Seungju Han$^{\spadesuit}$ \quad
Jack Hessel$^{\heartsuit}$ \quad 
Nouha Dziri$^{\heartsuit}$ \quad
Yejin Choi$^{\heartsuit\diamondsuit}$ \quad
Youngjae Yu$^{\heartsuit\clubsuit}$
\\
\small{
\small{$\spadesuit$ Seoul National University} \quad
$\heartsuit$ Allen Institute for Artificial Intelligence} \quad
\small{$\diamondsuit$ University of Washington} \quad
\small{$\clubsuit$ Yonsei University} \quad
\\
\texttt{wade3han@snu.ac.kr}
}

\ificcvfinal\thispagestyle{empty}\fi
\twocolumn[{
\renewcommand\twocolumn[1][]{#1}
\maketitle
\centering
\includegraphics[width=1.0\linewidth]{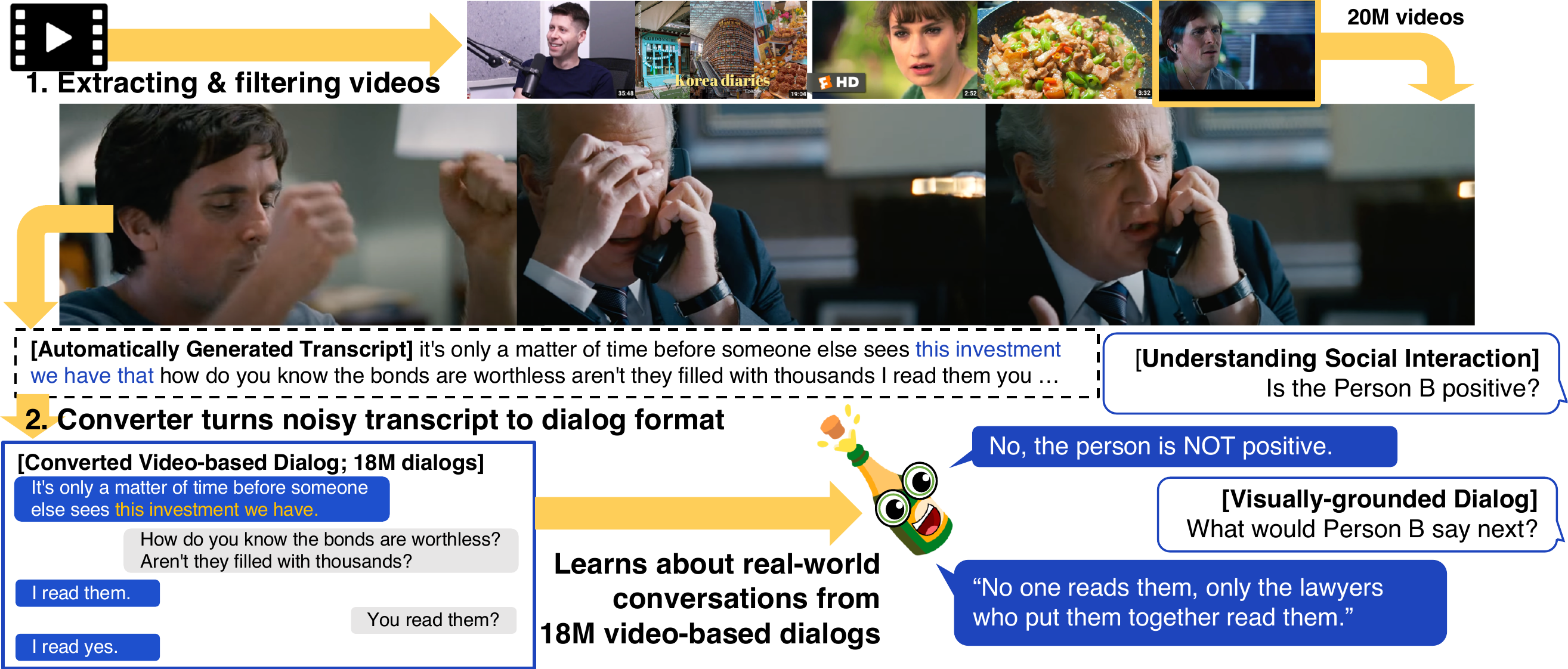}
\captionof{figure}{
\modelName is a generative model of real-world conversational frames trained on \datasetName, a dataset of 18M video-based dialogues.
\datasetNameNoEmoji is derived from public videos and their associated transcripts; a language model automatically grounds the conversation turns to the speakers.
\modelNameNoEmoji attains state-of-the-art results on four competitive vision-language benchmarks after fine-tuning.}
\label{fig:dataset_collection}
\vspace{5mm}
}]

\begin{abstract}
Visual information is central to conversation: body gestures and physical behaviour, for example, contribute to meaning that transcends words alone.
To date, however, most neural conversational models are limited to just text.
We introduce \modelName, a generative model of conversations that can account for visual contexts.
To train \modelNameNoEmoji, we collect and release \datasetName, a large-scale corpus of 18M video-based dialogues.
\datasetNameNoEmoji is constructed from web videos: crucial to our data collection pipeline is a pretrained language model that converts error-prone automatic transcripts to a cleaner dialogue format while maintaining meaning.

Human evaluation reveals that \datasetNameNoEmoji is more sensible and specific than prior resources (MMDialog \cite{feng2022mmdialog}, 1M dialogues), while maintaining visual-groundedness.
Experiments demonstrate that 1) \modelNameNoEmoji learns to conduct conversation from \datasetNameNoEmoji; and 2) when fine-tuned, it achieves state-of-the-art results on four vision-language tasks focused on real-world conversations.
We release data, models, and code at \href{https://seungjuhan.me/champagne}{https://seungjuhan.me/champagne}.
\end{abstract}

\section{Introduction}\label{sec:introduction}
\begin{figure*}[t]
\centering
\includegraphics[width=1.0\textwidth]{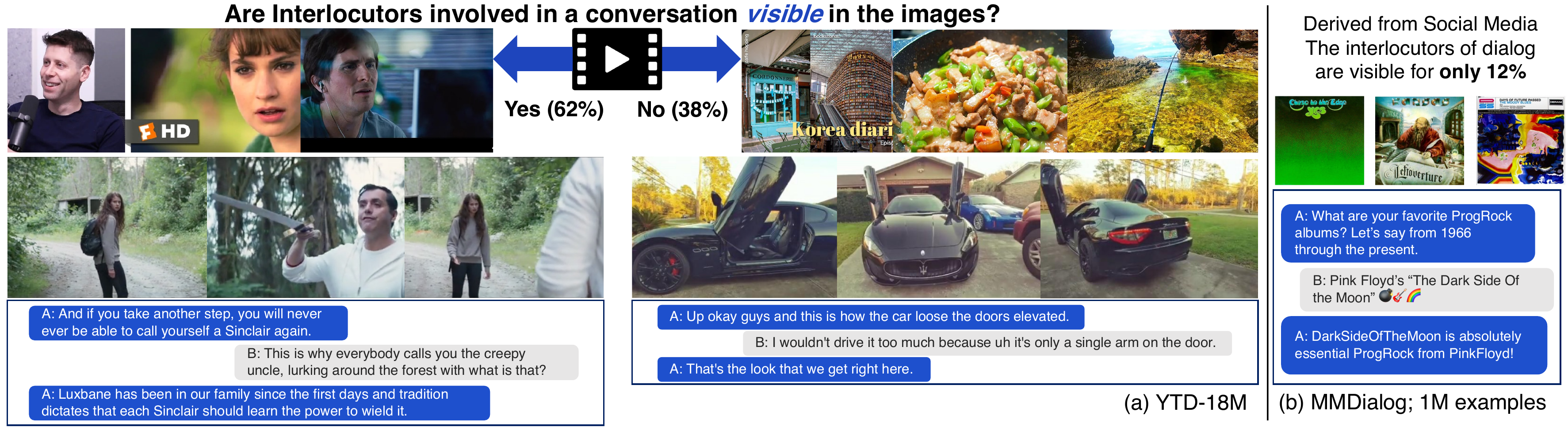}
\vspace{-5mm}
\caption{\small
Sample dialogues from two different datasets: (a) \datasetNameNoEmoji, and (b) the example from MMDialog presented in their paper~\cite{feng2022mmdialog}.
According to the human evaluation results, our \datasetNameNoEmoji dataset has a good balance of both social interactions (left; 62\%) and visually-grounded dialogue (right; 38\%), in contrast, MMDialog only has 12\% social interactions.
In addition, MMDialog is derived from social media, resulting in examples that contain emojis or other social media-specific elements, while \datasetNameNoEmoji is derived from videos and thus genuinely capture real-life communications.
}
\vspace{-5mm}
\label{fig:ytdialog_examples}
\end{figure*}
Conversation often relies on non-verbal cues:
visual information like physical expressions, body gesture, or the surrounding environment
are used by interlocutors to shape and understand meaning.
Figure~\ref{fig:dataset_collection}:
the two conversation participants appear stressed (in the first image: one person strikes the desk with his fists; in the second, the other person is rubbing his face); but the tension is not apparent simply from the transcript, which refers mostly to financial topics.
Other visual information suggests a broader context as well: that the conversation is taking place over the phone, that one person is older, and that one person is more formally dressed are all potentially important factors for understanding the conversation's meaning, yet none are reflected in the transcript. Indeed, visual perception provides information that can help machines understand the world in ways that text alone cannot~\cite{bisk2020experience, lecun2022path, cui2021empathic}.

In this paper, we propose \modelName\footnote{\textbf{C}onvers\textbf{A}tional \textbf{M}ultimodal \textbf{P}rompted \textbf{G}e\textbf{NE}rator.}, a generative model of conversations that learns from a large-scale video corpus.
\modelNameNoEmoji takes in video frames, 
a video title, and a dialogue context as input and returns a dialogue response as output.
The model learns from videos about two \emph{conversational frames:} 
1) \textit{Social Interaction}, where the conversation is observed from a 3rd-person perspective (\textit{e.g.} movies or interviews; Figure~\ref{fig:ytdialog_examples}, (a)-left); and 
2) \textit{Visually-grounded Dialogue}, where the conversation is observed from an embodied, first-person perspective (\textit{e.g.} ego-centric videos or chit-chatting through messenger applications; Figure~\ref{fig:ytdialog_examples}, (a)-right)).

To support training \modelNameNoEmoji, we collect and release a large-scale dataset, \datasetName, which is the largest publicly available dataset for real-world conversation learning.
\datasetNameNoEmoji is constructed from 20M YouTube videos:
we use a language model to convert the noisy transcripts automatically generated by YouTube into well-formatted dialogues associated with video frames.
Human evaluation shows that \datasetNameNoEmoji covers both social interaction and visually-grounded dialogue frames in balance, and surpasses the prior resource in terms of quality and scale (see \S\ref{subsec:analysing_dataset}).

After training, we demonstrate that \modelNameNoEmoji models %
generate high-quality next-turn utterances that account for visual contexts. 
Then, we conduct fine-tuning experiments, finding that:
1) \modelNameNoEmoji exhibits strong performance on open-domain text-only conversation benchmarks (\S\ref{subsec:open_domain_conversation});
2) \modelNameNoEmoji outperforms existing SOTA models on two \emph{social interaction} understanding benchmarks: CMU-MOSEI~\cite{zadeh2018multimodal} and Visual Comet~\cite{park2020visualcomet} (\S\ref{subsec:understand_social_interactions});
3) \modelNameNoEmoji outperforms SOTA models on two \emph{visually-grounded dialogue} benchmarks: Visual Dialog~\cite{das2017visual} and Image Chat~\cite{shuster2018image} (\S\ref{subsec:visual_grounded_dialogues}); and
4) ablations confirm the importance of various components of \datasetNameNoEmoji (\S\ref{sec:ablation}), \textit{e.g.} video frames.

In summary, our main contributions are: 
\begin{enumerate}[leftmargin=*]
  \item \datasetName, a large-scale dataset that contains 18M video-based dialogue that covers real-world conversational frames derived from 20M web videos.
  \item \modelName, a generative model that learns about real-world conversations from \datasetNameNoEmoji without any manual annotation.
  \item Experiments and ablations that demonstrate learning from a large-scale video-based dialogue dataset improves model performance on various tasks related to conversation.
\end{enumerate}

We publicly release code, the \datasetName dataset, and \modelName model checkpoints to facilitate future research on understanding real-world conversations from a visually-grounded perspective.
\section{Related Work}\label{sec:related_work}

Although recent language modeling advancements have enabled machines to engage in conversation and comprehend dialogue similar to humans \cite{zhang2019dialogpt, roller2020recipes, chen2020multi, adiwardana2020towards, chen-yang-2021-simple, dziri2022faithdial, han2022understanding}, these efforts have largely been confined to textual contexts.
In our work, we consider three real-world conversational frames: social interaction, visually-grounded dialogue, and open-domain text conversation. Social interaction involves conversations that incorporate visual cues and are observed from an external viewpoint (\textit{i.e.}, third-person perspective). 
Examples of these conversations include speaker sentiment analysis in videos as proposed by CMU-MOSEI~\cite{zadeh2018multimodal}, and human-focused commonsense-based captioning as introduced by Visual Comet~\cite{park2020visualcomet}.
Visually-grounded dialogue, on the other hand, refers to conversations that involve visual contexts from the perspective of the agent, \textit{i.e.}, from the first-person point of view. 
Visual Dialog~\cite{das2017visual}, which involves answering questions about an image in a conversational setting, and Image Chat~\cite{shuster2018image}, a chit-chat grounded to the image, are examples of visually-grounded dialogue.
Finally, open-domain text conversation refers to conversations that rely solely on textual contexts, and there have been multiple endeavors to teach machines to engage in natural communication on a wide range of topics~\cite{smith2020can, dinan2020second}.

Text-only dialogue models have access to abundant training resources~\cite{kim2022soda, thoppilan2022lamda, dziri2022faithdial, shuster2022blenderbot,  baumgartner2020pushshift, dziri2019augmenting}, allowing them to generate natural responses based on conversational context. However, despite their striking conversational abilities they lack the capability to comprehend visual contexts. 
To address this shortcoming, several studies have proposed multi-modal dialogue models \cite{shuster2020multi, sun2022multimodal}.
Few large-scale datasets such as MMDialog~\cite{feng2022mmdialog} are available for conversations that are grounded on visual contexts, yet the scale of the dataset is quite smaller than those of text-only dialogue datasets, making it difficult to train such models at scale.

\vspace{-13pt}

\paragraph{Large Vision-Language Models.}
Recent studies in vision-language research have shown that scaling the size of the model in conjunction with the dataset size can significantly enhance model performance across a range of tasks \cite{radford2021learning, ramesh2021zero}. Several frameworks have been introduced, including Unified-IO \cite{lu2022unified} and OFA \cite{wang2022ofa}, which propose a unified sequence-to-sequence model for modeling vision-language tasks. These frameworks demonstrate that a single model trained on a broad range of tasks can perform well on various tasks. Additionally, BLIP \cite{li2022blip} presents a technique for pretraining models on noisy datasets for both vision-language understanding and generation tasks. The Flamingo model \cite{alayrac2022flamingo}, trained on a large amount of noisy web data, demonstrates that it can adapt to tasks with few examples. While these models provide a general vision-language model, our research aims to enhance conversation capability by focusing on specifically learning about conversations.

\section{Approach}\label{sec:approach}
\subsection{Dataset Collection: \datasetName}\label{subsec:dataset_collection}
In this section, we describe our pipeline to collect \datasetName, which is outlined in Figure~\ref{fig:dataset_collection}.

\vspace{-13pt}

\paragraph{Extracting and Filtering Videos.}
The data collection process starts with downloading public YouTube videos. Each video is associated with a user-provided title, metadata, and a transcript generated by the YouTube's internal Automatic Speech Recognition system. 
We then filter videos by applying several steps using the associated metadata, following the strategy from~\citep{zellers2021merlot, zellers2022merlot}.
In particular, we use Python \textit{cld3} library, which uses a neural network for language identification, to filter out videos whose transcripts have a probability of being English less than 80\%.
We also discard videos that do not contain visual variation (\textit{e.g.}, a video of podcast with a static thumbnail) or those lacking objects in the thumbnails according to an image classification model~\cite{sandler2018mobilenetv2}. These steps resulted in the extraction of 20M videos.

We further process these 20M videos to build video segments. First, we build a list of segments by iterating through each video with a 60-second sliding window. Second, we filter out segments which have transcripts containing less than 30 words: short transcripts are unlikely to contain multiple conversation turns. 
Finally, we filter out transcripts that contain unsafe topics to ensure that the model does not learn harmful language. We utilize Rewire API~\cite{rewire} to detect toxic content in the transcripts.
At the end of the filtering stage, we end up with 18M video segments. 

\vspace{-13pt}

\paragraph{Converting Noisy Transcripts into Dialogues.}
Although each video is equipped with its own transcript,
they are usually not suitable to train a video-based dialogue model directly.
For example, transcripts do not provide any delimitation between the interlocutors.  A naive approach to address this problem is to use speaker diarization systems to determine ``who spoke when"~\citep{tranter2003investigation}. However, these systems suffer from low accuracy~\citep{park2022review}, thus perform poorly when turning a sequence of words to a well-structured dialogue. 

Instead, %
we train a \textit{converter} model to transform noisy transcripts into structured dialogues. Motivated by the in-context learning capabilities of GPT-3 models~\citep{brown2020language}, we prompt GPT-3 with few-shot examples and ask the model to generate well-formatted dialogues given noisy transcripts. 
However, using GPT-3 to process millions of samples is computationally expensive. Therefore, we train a smaller model using denoised transcripts sampled from GPT-3.
That is, we collect 40K input-output pairs from GPT-3 where the input is a noisy transcript and the output is the converted dialogue. Our \textit{converter} is a Unified-IO~\citep{lu2022unified} Base (241M params) fine-tuned on the generated pairs. We exclude video segments that have more than 150 words as the model cannot handle excessively long inputs. The pairs are divided into train and validation sets at a ratio of 99:1. The \textit{converter} achieves a high accuracy of 90.1\% on the validation set when using teacher-forcing to predict tokens,
suggesting high-quality dialogue generation given noisy transcripts (\textit{e.g.,} in Figure~\ref{fig:dataset_collection}, Step 2). 

\vspace{-13pt}

\paragraph{Converting Videos into Video-based Dialogues.}
To ensure that each utterance in the obtained dialogue matches correctly with the corresponding video frames, we employ Dynamic Time Warping~\cite{muller2007dynamic} to align the dialogues with the original noisy transcripts.
After the alignment, we use the timing information in the original noisy transcripts to estimate the start time of each dialogue turn, which in turn helps extract the corresponding video frame and minimize alignment errors caused by the conversion process.
Figure~\ref{fig:ytdialog_examples} shows some examples from \datasetNameNoEmoji. More details about collecting dataset can be found in Appendix~\ref{sec:appendix_dataset_collection}.

\subsection{Dataset Analysis}\label{subsec:analysing_dataset}
\begin{table}[t]
\centering
\footnotesize
\setlength{\tabcolsep}{3.0pt} 
\begin{tabular}{lll}
\toprule
Dataset & \datasetNameNoEmoji & MMDialog  \\
\midrule
Number of Dialog & \textbf{18M}  & 1M \\
\midrule
Is the interlocutors of dialog \textit{visible}? \\
~~~~Visible & \textbf{61.6\%$^*$} & 11.5\% \\
If \textbf{NOT} visible, then \\
~~~~how \textit{related} the dialog is to the image? & \textbf{2.589} & 2.580\\
~~~~how \textit{grounded} the dialog is to the image? & 2.429 & \textbf{2.463} \\
\midrule 
How \textit{sensible} is the dialog? & \textbf{2.495} & 2.479  \\
How \textit{specific} is the dialog? & \textbf{2.650$^*$} & 2.464 \\
Is the data containing \textit{explicit} content? \\
~~~~Sexually Explicit & \textbf{0.5\%$^*$} & 1.6\% \\
~~~~Hatespeech & \textbf{0.3\%$^*$} & 2.5\% \\
~~~~Others & \textbf{0.3\%$^*$} & 0.9\% \\
\bottomrule
\end{tabular}
\vspace{-2mm}
\caption{\small
Human evaluation results on \datasetNameNoEmoji and MMDialog about the visual contexts and the quality of dialogues.
\datasetNameNoEmoji covers more cases of social interactions, and dialogues exhibit better sensibleness and specificity compared to MMDialog.
$^*$ denotes statistically significance after independent two-sample t-test ($p < 0.5$).
Full breakdown of numbers are in Appendix~\ref{subsec:appendix_human_evaluation}.
}
\vspace{-5mm}
\label{tab:dataset_analysis}
\end{table}

To better understand our dataset, we conduct a human evaluation study comparing \datasetNameNoEmoji with MMDialog~\cite{feng2022mmdialog} --- the largest dataset for visually-grounded dialogue.
MMDialog is a dataset sourced from people's interaction in social media.
We randomly sample 500 examples from each dataset and ask three workers to assess each example for several factors.
For more information regarding the human evaluation, please refer to the Appendix~\ref{subsec:appendix_human_evaluation}.

\vspace{-13pt}
\paragraph{Dataset quality.}
To compare the dialogue quality, we ask workers to assess examples using two criteria: sensibleness and specificity~\cite{adiwardana2020towards}.
To be sensible, a dialogue should be reasonable, logical, and not confusing.
Specific dialogue is one that is not dull or generic.
Each human annotators rate the dialogue in specific aspect on a 3-point Likert scale, \textit{e.g.} for sensibleness "1" being "Not Sensible" and "3" being "Sensible".
Evaluation results are shown in Table~\ref{tab:dataset_analysis}. On average, \datasetNameNoEmoji received higher scores than MMDialog across the axes of sensibleness and specificity.
We suspect that MMDialog, which is derived from social media, may lack a natural conversation flow due to the non-consecutive nature of social media interactions.

\vspace{-13pt}
\paragraph{Social Interaction.}
Given the prevalence of third-person point of view in web videos,
we postulate that a substantial proportion of such videos feature social interactions. To investigate this claim, we enlisted the aid of workers to determine whether the conversation's interlocutors were visible in the image frames and, if so, whether their body language was present. Our findings, presented in Table~\ref{tab:dataset_analysis}, indicate that our video-based dialogue dataset has a considerably higher proportion of visible interlocutors (61.6\%) than MMDialog (11.5\%). This discrepancy can be explained by the fact that it is uncommon for interlocutors in social media interactions to reveal their identities through images. Moreover, when interlocutors are visible, workers accurately identified facial expressions in 83.6\% of cases, and tagged body posture in 64.7\% of them. These results suggest that our dataset presents a valuable resource for exploring body language in communication.

\vspace{-13pt}

\paragraph{Visual Grounding.}
Real-life conversations do not always have a direct relationship or grounding to images. Therefore, a higher degree of relevance between dialogues and images does not necessarily imply a higher quality dataset, although it offers more visual grounding opportunities for models to learn from. We ask workers to assess the degree of grounding between conversations and images on a 3-point Likert scale when interlocutors were not visible in the images. According to Table~\ref{tab:dataset_analysis}, \datasetNameNoEmoji exhibits grounding scores that are comparable to those of MMDialog, implying that models can acquire visual grounding knowledge from \datasetNameNoEmoji.

\vspace{-13pt}

\paragraph{Distribution of Visual Contexts.}
\begin{figure}[t]
\centering
\includegraphics[width=0.4\textwidth]{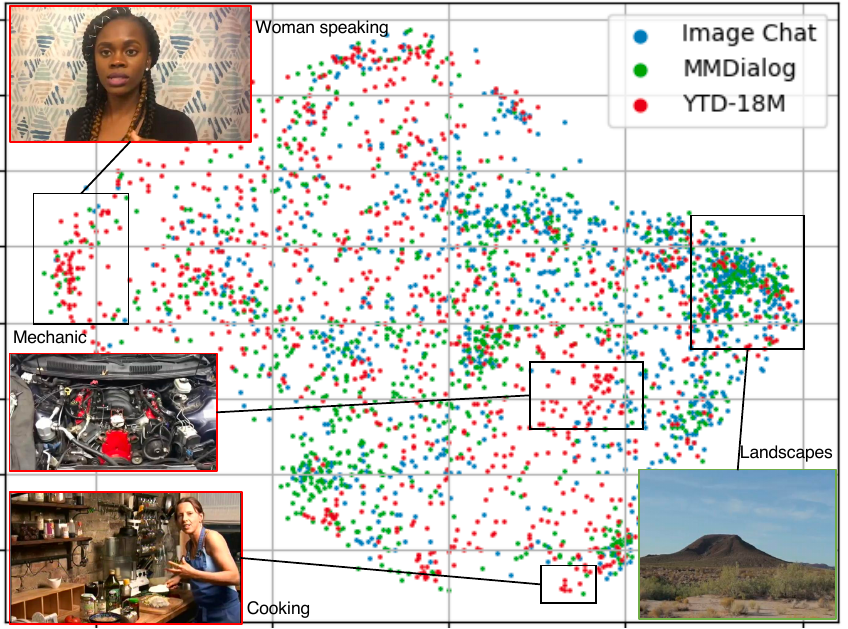}
\vspace{-2mm}
\caption{\small
Visual feature distributions of \datasetNameNoEmoji and other visually-grounded dialogue datasets. 
Our \datasetNameNoEmoji includes a wide range of visual contexts, with a particular emphasis on frames in which a person is speaking, in contrast to the other datasets (shown in the upper left cluster).}
\vspace{-5mm}
\label{fig:ytdialogue_visualize}
\end{figure}
In order to gain a better understanding of the visual representations encoded in \datasetNameNoEmoji from videos, we examine the distributions of visual features across three distinct datasets: Image Chat, MMDialog, and \datasetNameNoEmoji, that are grounded in dialogue. 
Similar to~\cite{liu2021visually}, we use CLIP ViT-L14~\cite{radford2021learning} trained on LAION-2B~\cite{schuhmannlaion}.\footnote{\cite{liu2021visually} conducted a similar study, but using ResNet50 ImageNet features.}%
We utilize this model to extract embeddings, and the embeddings are then projected into a 2D feature space by way of UMAP~\cite{mcinnes2018umap}. In order to conduct our analysis, we randomly sample 1K images from each of the aforementioned datasets.

Figure~\ref{fig:ytdialogue_visualize} illustrates the results, indicating that the datasets share some similarities but differ in certain aspects.  Notably, \datasetNameNoEmoji has a distinct distribution pattern from Image Chat and MMDialog, with a higher proportion of images featuring a person speaking, which is consistent with the previous discovery that \datasetNameNoEmoji  emphasizes social interactions. Additionally, \datasetNameNoEmoji  encompasses a broad range of diverse and specific topics, such as cooking or mechanics. Further information regarding the analysis of visual contexts can be found in the Appendix~\ref{sec:appendix_data_analysis}.

\vspace{-13pt}

\paragraph{Content Safety.}
We ask workers to identify whether the dialogues or images contain any potentially unsafe content, like sexually explicit material or hatespeech. As indicated in Table~\ref{tab:dataset_analysis}, our \datasetNameNoEmoji has fewer occurrences of sexually explicit content and hate speech when compared to MMDialog. We suspect that this discrepancy is due to the inclusion of safety filtering step in the data collection process for \datasetNameNoEmoji.

\vspace{-13pt}

\paragraph{Video Title as an Additional Feature.}
We further ask the workers to rate the relevance of the video title to the dialogue on a 3-point Likert scale, ranging from 1 (Not Related) to 3 (Related). The results
 show that 65\% of the videos have a title that is relevant to the dialogues, and 21\% have titles that were somewhat related. This led us to utilize the video titles as prompts when training our model. Our qualitative findings (\S\ref{subsec:open_domain_conversation}) suggest that \modelNameNoEmoji can be conditioned effectively by such prompts.

\subsection{Model: \modelName}\label{subsec:model}
\begin{figure}[t]
\centering
\includegraphics[width=0.35\textwidth]{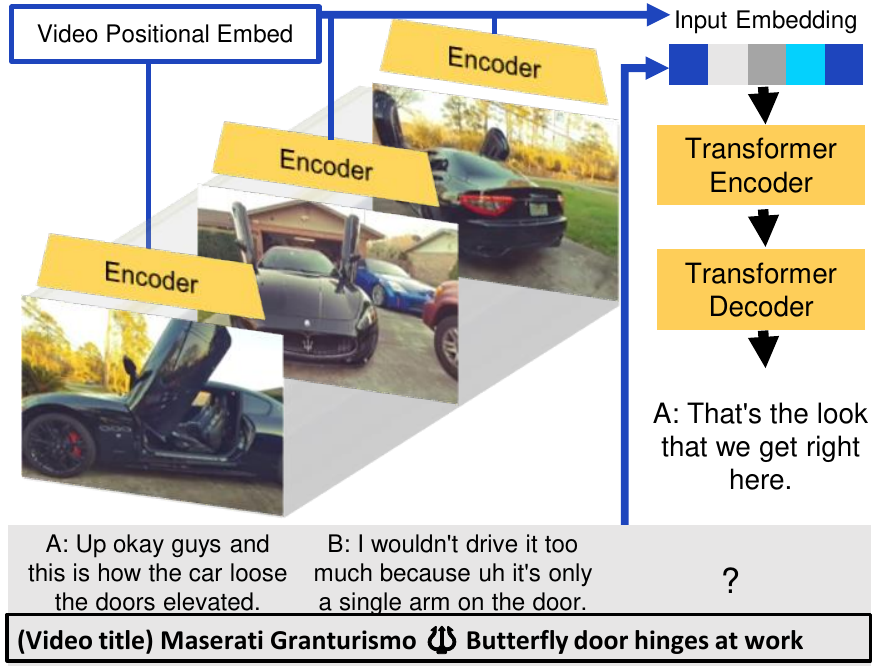}
\vspace{-2mm}
\caption{\small Training \modelName on \datasetName. The model takes a video title, a dialogue context, and image frames as an input and learns to predict the response.}
\vspace{-5mm}
\label{fig:model_training}
\end{figure}
\modelNameNoEmoji is a multimodal conversational agent that takes image frames, a prompt, and a dialogue context as an input and generates a response. 
The overview of training process is described in Figure~\ref{fig:model_training}.
The collected YouTube titles for each video serve as prompts and are denoted as $P$. 
Finally, the vision-based dialogue is denoted as $D = (P, I_1, T_1, ..., I_n, T_n)$, where $I_i$ and $T_i$ denote an image frame and the dialogue turn, respectively. 

\vspace{-13pt}

\paragraph{Architecture.}
The architecture used in \modelNameNoEmoji is based on the Unified-IO model proposed by \cite{lu2022unified}. This model is designed for vision-and-language tasks and operates as a sequence-to-sequence model. Although it can handle image and text inputs together, it is unable to handle multiple images. To address this limitation, we introduce \textit{video position embeddings} that can be learned and incorporated into \modelNameNoEmoji. Specifically, \modelNameNoEmoji  converts each frame of the video into a sequence of patch encodings using a visual encoder. It then adds video position embeddings to the patch encodings to capture temporal information of the video frames. The patch encodings from multiple image frames are averaged through mean pooling and fed into a Transformer encoder. Our experiments use three image frames per dialogue to train \modelNameNoEmoji.

\vspace{-13pt}

\paragraph{Training.}
We initialize \modelNameNoEmoji training with the pretrained weight from Unified-IO model that is pretrained on collection of C4~\cite{raffel2020exploring}, Wikipedia, ImageNet21K~\cite{ridnik2021imagenet}, and YFCC15M~\cite{radford2021learning} with a denoising objective.
Unified-IO model is trained in two stages, pre-training and the multi-task stage, and the weights from the pre-training stage, referred to as Unified-IO$_{PT}$\footnote{We run a pilot study and find that the model initialized from Unified-IO$_{PT}$ weights perform better on downstream tasks compared to the model initialized from Unified-IO.}, are used to initialize the \modelNameNoEmoji model.
We train the model using a next token prediction objective, which aims to maximize the likelihood of the target response $T_k$ when taking multiple images $I_{i\leq k}$, dialogue context $T_{i < k}$, and the video title $P$ as an input, \textit{i.e.}, $p_\theta(T_k|I_1, T_1, ... , I_{k-1}, T_{k-1}, T_k, P)$.

We present three versions of the model, \textsc{Base} (241M), \textsc{Large} (776M), and \textsc{XL} (2.9B).
We train models for 3 epochs on \datasetNameNoEmoji, and training \modelNameNoEmojiXL takes approximately 3 days on TPU v3-256 on Google Cloud Virtual Machines with T5X framework~\cite{roberts2022scaling}.
More details about hyperparameters are in Appendix~\ref{subsec:hparams}.

\begin{figure}[t]
\centering
\includegraphics[width=0.35\textwidth]{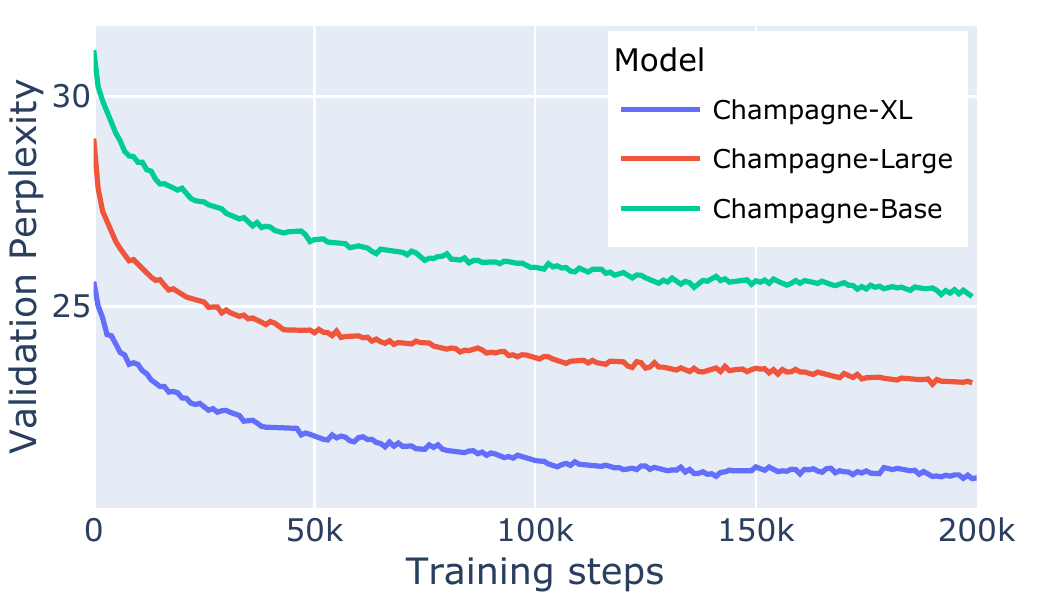}
\vspace{-3mm}
\caption{\small Validation perplexities during the training process of \modelNameNoEmoji models on \datasetNameNoEmoji.}
\vspace{-5mm}
\label{fig:training_curves}
\end{figure}
\begin{figure*}[t]
\centering
\includegraphics[width=0.85\textwidth]{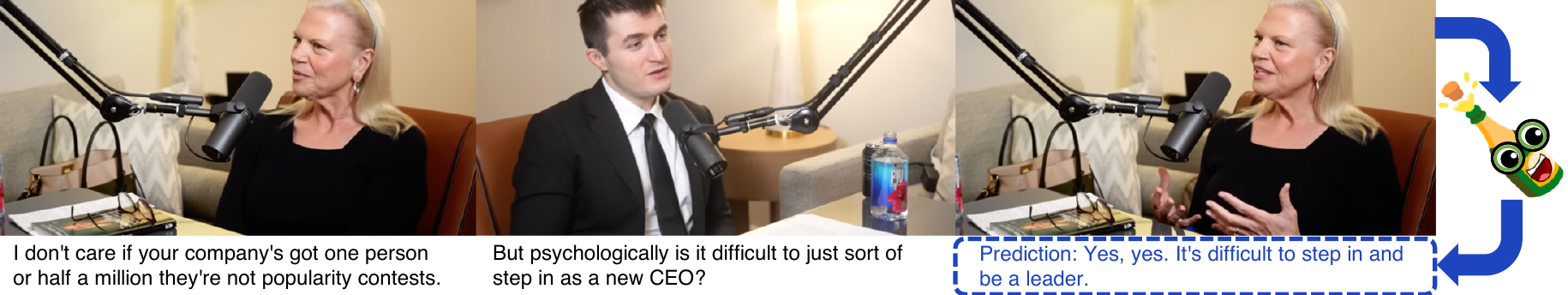}
\vspace{-2mm}
\caption{\small \modelNameNoEmojiXL predicting next utterance when the visual and dialogue contexts are given from the unseen video.}
\vspace{-1mm}
\label{fig:video_prediction}
\end{figure*}
\begin{table*}[t]
\centering
\footnotesize
\setlength{\tabcolsep}{5.0pt} 
\begin{tabular}{lcccccccccc|cccc}
\toprule
Dataset   & \multicolumn{2}{c}{BST} & \multicolumn{2}{c}{ConvAI2} & \multicolumn{2}{c}{ED} & \multicolumn{2}{c}{WOW} & \multicolumn{2}{c|}{WOI} & \multicolumn{2}{c}{IC} & \multicolumn{2}{c}{IC First Turn}\\
\midrule
Metric (on Test Set) & PPL & Dist-3 & PPL & Dist-3 & PPL & Dist-3 & PPL & Dist-3 & PPL & Dist-3 & PPL & Dist-3 & PPL & Dist-3 \\ 
\midrule
\textit{Without Fine-tuning} & & & & & \\
\champagne\modelNameNoEmojiXL  & 34.1 & 0.417 &	34.2 & 0.371 &	32.8 & 0.544 &	31.2 & 0.705 &	34.8 & 0.641 &	50.5 & 0.213 & 69.9 & 0.135 \\
Reddit 2.7B~\cite{roller2020recipes} & 20.0 & 0.390 & 24.2 & 0.286 & 14.0 & 0.418 & 24.3 & 0.458 & 23.5 & 0.473 & 40.9 & 0.152 & 66.3 & 0.011  \\
\midrule
\textit{Fine-tuned} \\
Multimodal Blender 2.7B~\cite{shuster2020multi} & \textbf{14.6} & 0.418 & 13.8 & 0.300 & \textbf{11.4} & 0.306 & \textbf{15.6} & 0.527 &  19.8 & 0.522 & 24.0 & 0.200 & 25.9 & 0.193  \\
Unified-IO$_{PT}$ \textsc{Base} & 20.4 & 0.280 & 16.5 & 0.199 &	19.8 & 0.352 &	22.6 & 0.480 & 	26.0 & 0.449  & 29.9 & 0.106 & 29.6 & 0.113 \\
Unified-IO$_{PT}$ \textsc{Large} & 17.7 & 0.284 &	15.0 & 0.176 & 15.8 & 0.304 &	18.6 & 0.484 & 	21.3 & 0.436 & 	26.8 & 0.108 & 	26.7 & 0.096  \\
Unified-IO$_{PT}$ \textsc{XL} & 15.7 & 0.346 & 12.9 & 0.250 & 	14.9 & 0.393 & 	16.5 & 0.580 &	18.9 & 0.515 & 	20.5 & 0.182 & 22.2 & 0.163  \\
\champagne\modelNameNoEmojiBase & 19.1& 0.395 & 15.2  & 0.361 & 19.0 & 0.421 & 21.3 & 0.651 & 24.1 & 0.584 & 27.2 & 0.174 & 27.4 & 0.220  \\
\champagne\modelNameNoEmojiLarge  & 16.9	& 0.390 & 13.6 & 0.353 &	18.6 & 0.454 & 18.5 & 0.665 & 	21.7 & 0.590 & 22.9 & 0.205 & 23.6 & 0.256  \\
\champagne\modelNameNoEmojiXL   & 15.1 & \textbf{0.434}& 	\textbf{12.3} & \textbf{0.390}& 	16.2& \textbf{0.499} & 	16.1 & \textbf{0.690} & 	\textbf{18.5} & \textbf{0.624} &	\textbf{19.3} & \textbf{0.247} &	\textbf{21.2} & \textbf{0.278} \\
\bottomrule
\end{tabular}
\vspace{-2mm}
\caption {\small Automatic evaluation results on open-domain text conversation benchmarks and Image Chat (IC). 
For fair comparison, all perplexities (PPL, $\downarrow$) are normalized to be in the space of GPT-2 tokenizer~\cite{wolf-etal-2020-transformers}.
Dist-3 ($\uparrow$) is calculated over corpus-level (Inter Dist-3).
\textit{IC First Turn} indicates the situation in which the model generates the response given image but no textual context from IC, which is to highlight model's ability to ground on visual contexts.
Note that Multimodal Blender 2.7B is a fine-tuned version of Reddit 2.7B.
}
\vspace{-3mm}
\label{tab:open_domain_chat}
\end{table*}
\section{Experiments}\label{sec:experiments}
Figure~\ref{fig:training_curves} shows the training curves for \modelNameNoEmoji models that were trained on \datasetNameNoEmoji.
We evaluate the sensibleness and specificity of the response generated by \modelNameNoEmoji models by computing perplexities~\cite{jelinek1977perplexity} on the validation set of \datasetNameNoEmoji. Perplexity (PPL) has been shown to be a reliable indicator of these qualities~\cite{adiwardana2020towards}. The training curves in Figure~\ref{fig:training_curves} demonstrate that the models are capable of learning to generate appropriate next-turn utterances given visual and dialogue contexts, and that they generalize well to the validation set when trained on \datasetNameNoEmoji. A qualitative example of this successful generalization can be seen in Figure~\ref{fig:video_prediction}.

We next conduct experiments on a range of benchmarks that assess the model's performance on real-world conversation tasks. Additional information about the benchmarks and the evaluation metrics used can be found in Appendix~\ref{sec:appendix_datasets}.

\subsection{Open-domain Text Conversation}
\label{subsec:open_domain_conversation}
To investigate whether \modelNameNoEmoji models are capable of learning chit-chat skills from \datasetNameNoEmoji, we conduct automatic evaluations on open-domain text conversation benchmarks. These benchmarks assess the model's ability to produce appropriate responses given a textual dialogue context in a conversational scenario. Specifically, we evaluate the model's performance on five text-only benchmarks: BST~\cite{smith2020can}, ConvAI2~\cite{dinan2020second}, ED~\cite{rashkin2018towards}, WOW~\cite{dinan2018wizard}, and WOI~\cite{komeili2021internet}. We employ two metrics to evaluate the models: PPL and Dist-3~\cite{li2016diversity}. Dist-3 is a metric that has been shown to reflect the diversity and interestingness of the response~\cite{Han2022MeasuringAI}. Results are shown in Table~\ref{tab:open_domain_chat} (left).

Prior to fine-tuning, both Unified-IO$_{PT}$ \textsc{XL} and Unified-IO \textsc{XL} exhibited perplexities over 100, which suggests that these models generate unintelligible sentences. In contrast, \modelNameNoEmojiXL achieved significantly lower perplexities on all benchmarks, indicating that the model has effectively learned chit-chatting capabilities from \datasetNameNoEmoji.
Next, we fine-tune both Unified-IO$_{PT}$ and \modelNameNoEmoji on a mixture of dialogue benchmarks, with the multi-task training weight set to BST: ConvAI2: ED: WOW: WOI: IC $= 1:3:3:3:3:3$, following the approach described in~\cite{roller2020recipes}. After fine-tuning, \modelNameNoEmoji achieves significantly higher Dist-3 scores and lower PPL values than the same-sized Unified-IO$_{PT}$, indicating that \modelNameNoEmoji has successfully learned about textual conversations from \datasetNameNoEmoji. Our experimental results further suggest that larger models tend to perform better overall, with \modelNameNoEmojiXL reporting Dist-3 scores that surpassed those of other baselines on all benchmarks, demonstrating its ability to generate diverse and interesting responses. However, in terms of PPL, \modelNameNoEmojiXL shows weaker performance on BST, ED, and WOW compared to the Multimodal Blender 2.7B. We conjecture that this may be due to the influence of the large text-based dialogue dataset (the Pushshift dataset~\cite{baumgartner2020pushshift}), which contains 3B Reddit comments. Indeed, the Reddit 2.7B model, the text-only model which is trained on the Pushshift dataset, performs better than \modelNameNoEmojiXL without fine-tuning, suggesting that the Pushshift dataset has a distribution that is more similar to those of these benchmarks than \datasetNameNoEmoji. 
We expect that further improving the model's performance on these benchmarks may require training \modelNameNoEmojiXL on the Pushshift dataset in addition to fine-tuning it on these benchmarks.

\subsection{Understanding Social Interactions}\label{subsec:understand_social_interactions}
\begin{table}[t]
\centering
\footnotesize
\begin{tabular}{lcc}
\midrule
Metric & Acc. (\%) ($\uparrow$) & F1 ($\uparrow$) \\
\midrule
UniMSE~\cite{hu2022unimse} & 85.8 & 0.858 \\
MAG-BERT~\cite{rahman-etal-2020-integrating} & 84.7 & 0.845 \\
Unified-IO$_{PT}$ \textsc{Base} & 70.1 & 0.780\\
Unified-IO$_{PT}$ \textsc{Large} & 72.9 & 0.812\\
Unified-IO$_{PT}$ \textsc{XL} & 81.5 & 0.869\\
\midrule
\champagne\modelNameNoEmojiBase & 82.3 & 0.879 \\
\champagne\modelNameNoEmojiLarge & 83.9 & 0.892\\
\champagne\modelNameNoEmojiXL  & \textbf{86.1} & \textbf{0.899}\\
\bottomrule
\end{tabular}
\vspace{-2mm}
\caption {\small
Evaluation results on test split of CMU-MOSEI sentiment analysis task. 
}
\vspace{-3mm}
\label{tab:mosei}
\end{table}
\begin{table}[t]
\centering
\footnotesize
\setlength{\tabcolsep}{6pt} 
\begin{tabular}{lcccc}
\toprule
Metric & \multicolumn{2}{c}{\makecell{CIDEr-D ($\uparrow$)}} & \multicolumn{2}{c}{\makecell{BLEU-4 ($\uparrow$)}} \\
\midrule
Split & Val & Test & Val & Test \\
\midrule
Unified-IO XL & 0.212 & - &  0.073 & - \\
SOTA on Public Leaderboard~\cite{visualcomet} & - & 0.184 & - & 0.041 \\
Unified-IO$_{PT}$ \textsc{Base} & 0.344 & - & 0.095 & -\\
Unified-IO$_{PT}$ \textsc{Large} & 0.356 & - & 0.098 & -\\
Unified-IO$_{PT}$ \textsc{XL} & 0.384 & - & 0.105 & -\\
\midrule
\champagne\modelNameNoEmojiBase & 0.351 & - & 0.097 & -\\
\champagne\modelNameNoEmojiLarge & 0.369 & 0.322 & 0.102 & 0.080 \\
\champagne\modelNameNoEmojiXL & \textbf{0.399} & \textbf{0.354} & \textbf{0.109} & \textbf{0.092}  \\
\bottomrule
\end{tabular}
\vspace{-2mm}
\caption {\small
Evaluation results on Visual Comet.
}
\vspace{-5mm}
\label{tab:visualcomet}
\end{table}
We evaluate the performance of three variants of \modelNameNoEmoji and Unified-IO$_{PT}$ on two benchmarks that measure social interactions: the sentiment analysis task in CMU-MOSEI~\cite{zadeh2018multimodal} and Visual Comet~\cite{park2020visualcomet}.
Table~\ref{tab:mosei} and Table~\ref{tab:visualcomet} show the results on CMU-MOSEI and Visual Comet, respectively.
Results show that fine-tuning \modelNameNoEmojiXL on these benchmarks leads to the highest accuracy and F1 score among the baselines, including the state-of-the-art model (UniMSE) on CMU-MOSEI. Additionally, our model outperforms the state-of-the-art result on Visual Comet, both on the validation and test sets.
Furthermore, the models fine-tuned from \modelNameNoEmoji consistently demonstrate significant improvements compared to the models fine-tuned from same-sized Unified-IO$_{PT}$ on both CMU-MOSEI and Visual Comet. This indicates that our model effectively learns meaningful representations about social interactions from \datasetNameNoEmoji.

\subsection{Visually-grounded Dialogues}\label{subsec:visual_grounded_dialogues}
\begin{figure*}[t]
\centering
\includegraphics[width=0.8\textwidth]{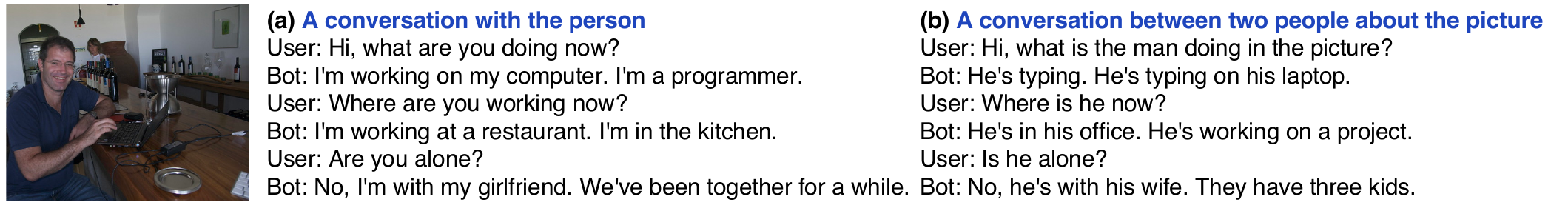}
\vspace{-3mm}
\caption{
\small Examples of conversation based on unseen images from COCO~\cite{lin2014microsoft} and prompts (colored in blue) between person and \modelNameNoEmojiXL finetuned on mixture of dialogue benchmarks.
The utterances with the prefix \textit{Bot:} are generated by the model.
(a) represents a conversational frame of social interaction (bot pretends to be the person in the picture), while (b) represents a conversational frame of visually-grounded dialogue (bot talks about the person in the picture).
}
\vspace{-3mm}
\label{fig:inference_results}
\end{figure*}
\begin{table}[t]
\centering
\footnotesize
\begin{tabular}{lc}
\toprule
Metric & NDCG ($\times$100) ($\uparrow$) \\
\midrule
\textit{Zero-shot} \\
Flamingo-80B & 52.0 \\
ESPER~\cite{yu2022multimodal} & 22.3 \\
FROMAGe~\cite{koh2023grounding} & 16.5 \\
\champagne\modelNameNoEmojiXL & 25.5 \\
\midrule
\textit{Fine-tuned} \\
Flamingo-80B & 61.8 \\
AlignVD~\cite{chen2022unsupervised} & 67.2 \\
Unified-IO$_{PT}$ \textsc{Base} & 58.9 \\
Unified-IO$_{PT}$ \textsc{Large} & 60.3 \\
Unified-IO$_{PT}$ \textsc{XL} & 65.6 \\
\champagne\modelNameNoEmojiBase & 60.0 \\
\champagne\modelNameNoEmojiLarge & 62.5 \\
\champagne\modelNameNoEmojiXL & \textbf{68.2} \\
\bottomrule
\end{tabular}
\vspace{-2mm}
\caption {\small
Evaluation results on Visual Dialog valid set in finetuned and zero-shot settings.
\modelNameNoEmojiXL shows the best zero-shot performance except for Flamingo-80B, which has 30$\times$ more parameters, and \modelNameNoEmojiXL achieves the state-of-the-art result on finetuned setting.
For fair comparison, we report baselines that do not use additional dense annotations to finetune the model.
}
\vspace{-4mm}
\label{tab:visdial}
\end{table}

\begin{table*}[t]
\centering
\footnotesize
\setlength{\tabcolsep}{3.2pt} 
\begin{tabular}{l|lllll|lllll}
\toprule
Dataset & \multicolumn{5}{c|}{Image Chat} & \multicolumn{5}{c}{Image Chat First Turn} \\
\midrule
 Metric & Sensible & Specific & Grounding & Style & Avg. &  Sensible & Specific & Grounding & Style & Avg. \\
\midrule
Multimodal Blender 2.7B & 81.3\% & 66.7\% & \textbf{98.7\%} & 69.0\% & 78.9\% & \textbf{93.3\%} & 70.0\% & 85.0\% & 82.3\% & 82.6\% \\
Unified-IO$_{PT}$ \textsc{XL} & \textbf{84.7\%} & 88.3\% & 98.0\% & 84.3\% & 88.8\% & 92.0\% & 83.0\% & 87.7\% & 82.0\% & 86.3\% \\
\champagne\modelNameNoEmojiXL & 82.7\% & \textbf{93.7\%$^{UM}$} & \textbf{98.7\%} & \textbf{90.0\%$^{UM}$} & \textbf{91.3\%$^{UM}$} & 92.0\% & \textbf{93.0\%$^{UM}$} & \textbf{90.1\%$^{M}$} & \textbf{88.3\%$^{UM}$} & \textbf{90.8\%$^{UM}$}\\
\bottomrule
\end{tabular}
\vspace{-2mm}
\caption {\small Human evaluation results on Image Chat and Image Chat First Turn.
\textit{Grounding} denotes how well the generated response is grounded to the image, \textit{Style} denotes how well the generated response reflects the given style, and \textit{Avg.} denotes the average percent of results.
We run independent two-sample t-tests, and we denote as $^M$ and $^U$ when \modelNameNoEmojiXL is statistically significant ($p < 0.05$) over Multimodal Blender 2.7B and Unified-IO$_{PT}$~\textsc{XL}, respectively.
} 
\vspace{-3mm}
\label{tab:open_domain_human_eval}
\end{table*}
We run experiments on two visually-grounded dialogue benchmarks: Visual Dialog~\cite{das2017visual} and Image Chat~\cite{shuster2018image}. Table~\ref{tab:visdial} presents the results of models on Visual Dialog in both zero-shot and fine-tuned settings.
Similarly, Table~\ref{tab:open_domain_chat} (right) shows the results of models fine-tuned on combination of dialogue datasets on Image Chat and Image Chat First Turn. 
Fine-tuning \modelNameNoEmoji on those benchmarks outperforms fine-tuning the same sized Unified-IO$_{PT}$ model on both tasks. This suggests that the model effectively learns from \datasetNameNoEmoji, and larger models tend to perform better. For Visual Dialog zero-shot setting, \modelNameNoEmojiXL outperforms ESPER and FROMAGe in terms of NDCG~\cite{jarvelin2002cumulated}. In the fine-tuned setting, \modelNameNoEmojiXL achieves the highest NDCG compared to Flamingo-80B and the state-of-the-art model (AlignVD). However, in the zero-shot setting, Flamingo-80B performs significantly better than \modelNameNoEmojiXL, which might be due to Flamingo's larger parameters and training data. Based on our scaling observations, we suspect that training Flamingo-80B or a similar sized model on \datasetNameNoEmoji could lead to further performance improvements. For Image Chat and Image Chat First Turn, \modelNameNoEmojiXL shows the lowest PPL and highest Dist-3 scores among the baselines.

We additionally carry out a human evaluation to compare our best model with the baselines trained on Image Chat. We randomly choose 100 examples from both Image Chat and Image Chat First Turn. For each example, we asked three workers to rate the dialogue responses generated by the model on four aspects: (1) sensibleness, (2) specificity, (3) grounding to the image, and (4) relevance to the given style. The workers were presented with positive/negative options for each aspect. For instance, for the grounding aspect, they were asked to choose between "Yes, the response is grounded to the image" or "No, the response is not grounded to the image." Our evaluation results, presented in Table~\ref{tab:open_domain_human_eval}, indicate that our model, \modelNameNoEmojiXL, outperforms the baselines in terms of specificity, grounding, and style. As for sensibleness, our model shows comparable scores to the best results. More details on the human evaluation can be found in Appendix~\ref{subsec:appendix_human_evaluation}.
Lastly, Figure~\ref{fig:inference_results} presents some sample conversations between a human and our model. The figure demonstrates that the model can engage in conversations based on images and can be prompted to respond accordingly.

\subsection{Ablations}\label{sec:ablation}
\begin{table}[t]
\centering
\footnotesize
\setlength{\tabcolsep}{1pt} 
\begin{tabular}{lccc}
\toprule
Dataset & \makecell{CMU-MOSEI} &  \multicolumn{2}{c}{\makecell{IC First Turn}} \\
\midrule
Metric & Acc. (\%) & PPL ($\downarrow$) & Dist-3 ($\uparrow$) \\
\midrule
\champagne\modelNameNoEmojiLarge trained on: \\
~~~~YTD-18M & \textbf{83.9} & \textbf{21.6} & \textbf{0.260} \\
~~~~YTD-2M & 83.2 & 28.3 & 0.190\\
~~~~YTD-2M w/ single video frame & 78.9 & 29.3 & 0.066\\
~~~~YTD-2M w/o video frames & 75.6 & 29.7 & 0.073 \\
~~~~YTD-2M w/o transcript to dialog & 74.9 & 29.0 & 0.143\\
\bottomrule
\end{tabular}
\vspace{-2mm}
\caption {\small Ablation study on the test set of CMU-MOSEI and Image Chat First Turn to validate the effect of \datasetNameNoEmoji.}
\vspace{-5mm}
\label{tab:ablation}
\end{table}
To investigate the impact of components in \datasetNameNoEmoji, we conduct ablation studies on CMU-MOSEI and Image Chat First Turn. Our evaluation of models on Image Chat First Turn only involves fine-tuning on Image Chat and excludes any text-only dialogue datasets.
\vspace{-13pt}

\paragraph{Number of Examples.}
The performance drop in both tasks is evident when reducing the number of examples in \datasetNameNoEmoji from 18M to 2M, as seen in Table~\ref{tab:ablation}. These findings are consistent with previous works~\cite{hoffmann2022training, brown2020language} and highlight the benefits of learning from a larger dataset. Notably, the significance of the performance drop is greater in the visually-grounded dialogue task of Image Chat, underscoring the crucial role of the number of examples in such tasks.

\vspace{-13pt}

\paragraph{Video Frames.}
 To validate whether the models learn from visual contexts, we conduct two experiments. The first experiment involves training the model without video frames, while the second experiment utilizes only a single video frame for training. Table~\ref{tab:ablation} presents the results of these experiments, which show a decrease in performance for both tasks. These findings support the idea that \datasetNameNoEmoji provides an opportunity for the model to learn visual grounding. Notably, the accuracy drop is more significant for CMU-MOSEI, where training with a single frame results in an accuracy drop from 83.2\% to 78.9\%, and training without video frames results in an even greater accuracy drop to 75.6\%. This suggests that relying on visual cues is essential for understanding sentiment in social interaction. Furthermore, in Image Chat, there is a substantial drop in Dist-3, indicating that models trained without visual contexts may generate generic responses that are irrelevant to the images.

\vspace{-13pt}

\paragraph{Dialogue Format.}
By comparing the model trained on the noisy transcript to the model trained on dialogue format, we aim to demonstrate how the latter can improve the machine's ability to learn representations of conversation. 
In Table~\ref{tab:ablation}, we observe a decrease in performance for both tasks.
Notably, on the CMU-MOSEI dataset, the accuracy drops from 83.2\% to 74.9\% without the dialogue conversion process. We speculate that the conversion process helps the dataset recover crucial speaker information that is missing from the noisy transcript. This information could prove beneficial for learning representations of social interaction.

\section{Conclusion}
We introduce \datasetName, a large-scale video-based dialogue dataset and \modelName, a model that learns about the real-world conversation from \datasetNameNoEmoji.
Our experiments show that \modelNameNoEmoji exhibits strong performance in various real-world conversation tasks, indicating that video-based dialogues can help models to learn about real-world conversations. 
We will release our data and model checkpoints for research purposes. 
In future work, we plan to explore the potential of utilizing auditory signals from videos to further enhance our understanding of real-world conversations.

\section{Ethical Considerations}
In this paper, we introduce a large-scale dataset derived from publicly available YouTube videos.
With an emphasis on teaching machines about real-world conversation, our dataset includes frames that present the interlocutors.
It might capture their facial expressions and body cues, however, may give rise to user privacy concerns.
To mitigate this issue, we release only the video IDs instead of raw videos, following the prior works~\cite{zellers2021merlot, zellers2022merlot}.
It allows users to delete their video from YouTube, thereby excluding them from the dataset.
To strengthen user privacy even further, future directions may include works such as anonymizing faces and speech from the videos, and deidentifying personal information from the transcripts.

{\small
\bibliographystyle{ieee_fullname}
\bibliography{iccv}
}

\clearpage
\appendix
\section{Details of Dataset Collection}\label{sec:appendix_dataset_collection}
\paragraph{Safety Filtering.}
We use Rewire API~\cite{rewire} to filter out unsafe contents from videos.
Rewire API identifies abusive, hateful, profane, violent, or sexually explicit content.
However, we have discovered that the API is not accurate enough to detect profanities and violent languages in video transcripts.
Thus, we only use API to detect abusive, hateful, or sexually explicit content.
We set thresholds of 0.99534, 0.83790, 0.99562 to filter out unsafe contents for abuse, hate, and sexually explicit labels, respectively.

\paragraph{Aligning Video and Dialogue.}
We use Dynamic Time Warping~\cite{muller2007dynamic} to align the dialogue (text) with the video frames.
In particular, we first calculate the distance between the noisy transcript and the converted dialogue using Levenshtein distance.
We then employ Dynamic Time Warping to align the words and minimize the distance between the transcript and the dialogue.
Following that, using the timing information associated with the transcripts, we estimate the start time of each utterances in the dialogue.
We extract the video frame using the start timing of the utterance, resulting in a video-based dialogue with video frames and the dialogue turns $(I_1, T_1, ..., I_n, T_n)$.

\section{Human Evaluation}\label{subsec:appendix_human_evaluation}

\begin{table}[t]
\centering
\footnotesize
\setlength{\tabcolsep}{5.0pt} 
\begin{tabular}{lrr}
\toprule
Dataset & \datasetNameNoEmoji & MMDialog  \\
\midrule
Number of Dialog & 18M  & 1M \\
\midrule 
\textbf{How \textit{sensible} is the dialog?} \\
~~~~Natural (3) & 60.1\% & 57.7\% \\
~~~~Slightly Natural (2) & 29.4\% & 32.6\% \\
~~~~Unnatural (1) & 10.5\% & 9.7\% \\
~~~~Avg. Score & \textbf{2.495} & 2.479  \\
\textbf{How \textit{specific} is the dialog?} \\
~~~~Specific (3) & 70.6\% & 59.7\% \\
~~~~Slightly Specific (2) & 21.3\% & 26.9\% \\
~~~~Unspecific (1) & 8.1\% & 13.3\%\\
~~~~Avg. Score & \textbf{2.650$^*$} & 2.464 \\
\textbf{Is the data containing \textit{explicit} content?} \\
~~~~Sexually Explicit & \textbf{0.5\%$^*$} & 1.6\% \\
~~~~Hatespeech & \textbf{0.3\%$^*$} & 2.5\% \\
~~~~Others & \textbf{0.3\%$^*$} & 0.9\% \\
\bottomrule
\end{tabular}
\caption{
Full breakdown for human evaluation results on \datasetNameNoEmoji and MMDialog about the quality of dialogues. 
$^*$ denotes statistically significance after independent two-sample t-test ($p < 0.5$).
}
\label{tab:dataset_analysis_full}
\end{table}

\begin{table}[t]
\centering
\footnotesize
\setlength{\tabcolsep}{5.0pt} 
\begin{tabular}{lrr}
\toprule
Dataset & \datasetNameNoEmoji & MMDialog  \\
\midrule
\textbf{Is the interlocutors of dialog \textit{visible}?} \\
~~~~Visible & \textbf{61.6\%$^*$} & 11.5\% \\
\textit{If \textbf{NOT} visible, then} \\
\textbf{is the dialog \textit{related} to the image(s)?} \\
~~~~Related (3) & 71.0\% & 67.5\% \\
~~~~Slightly Related (2) & 16.8\% & 22.9\%  \\
~~~~Unrelated (1) & 12.2\% & 9.6\% \\
~~~~Avg. Score & \textbf{2.589} & 2.580\\
\textbf{is the dialog \textit{grounded} to the image(s)?} \\
~~~~Grounded (3) & 62.3\% & 59.9\% \\
~~~~Slightly Grounded (2) & 18.2\% & 26.5\%\\
~~~~Not Grounded (1) & 19.4\% & 13.6\%  \\
~~~~Avg. Score & 2.429 & \textbf{2.463} \\
\bottomrule
\end{tabular}
\caption{
Full breakdown for human evaluation results on \datasetNameNoEmoji and MMDialog about visual contexts. 
$^*$ denotes statistically significance after independent two-sample t-test ($p < 0.5$).
}
\label{tab:dataset_analysis_visual_full}
\end{table}

\begin{figure*}[t]
\centering
\includegraphics[width=0.85\textwidth]{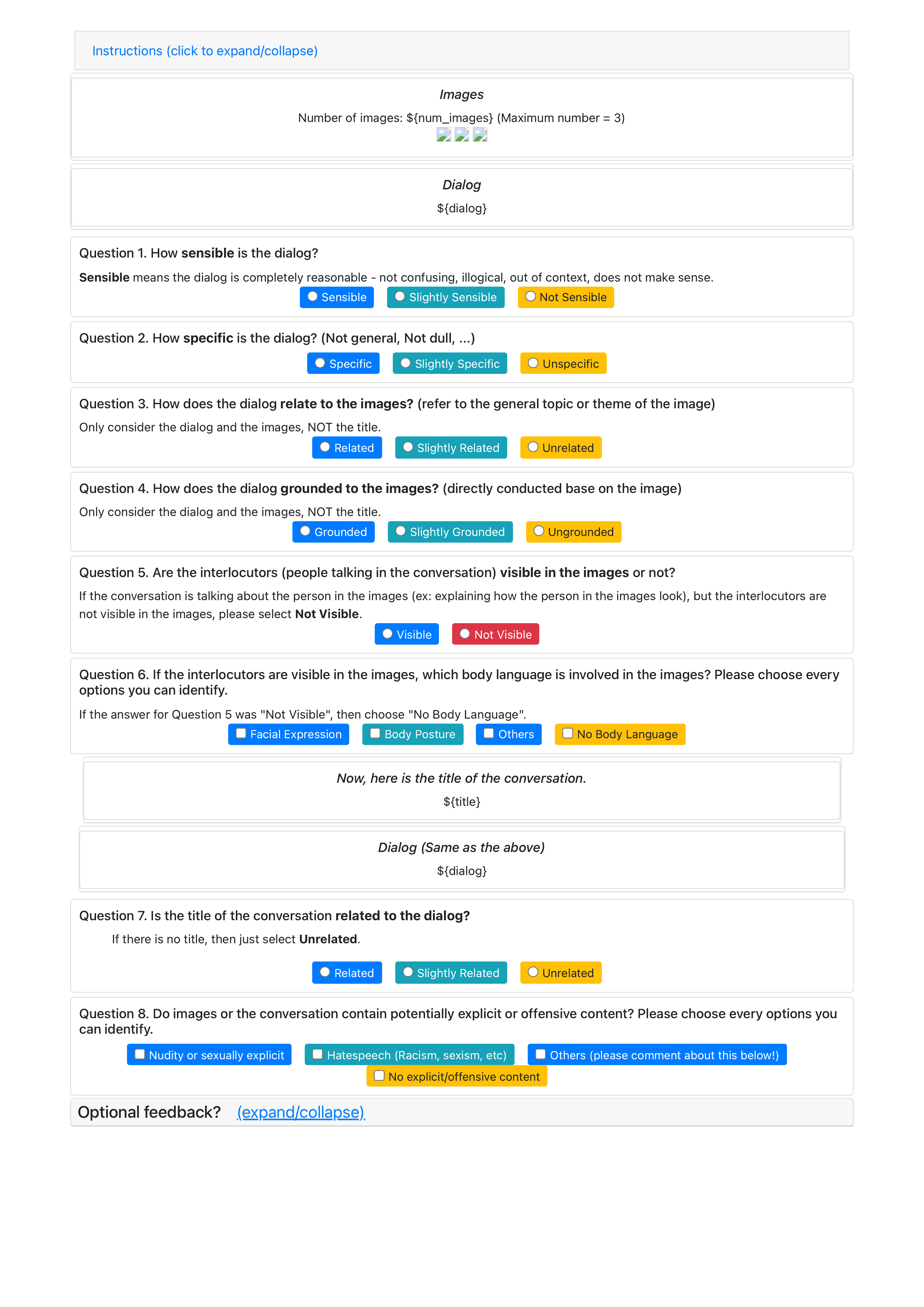}
\caption{An interface for evaluating datasets on Amazon Mechanical Turk.}
\label{fig:dataset_evaluation}
\end{figure*}
\begin{figure*}[t]
\centering
\includegraphics[width=0.85\textwidth]{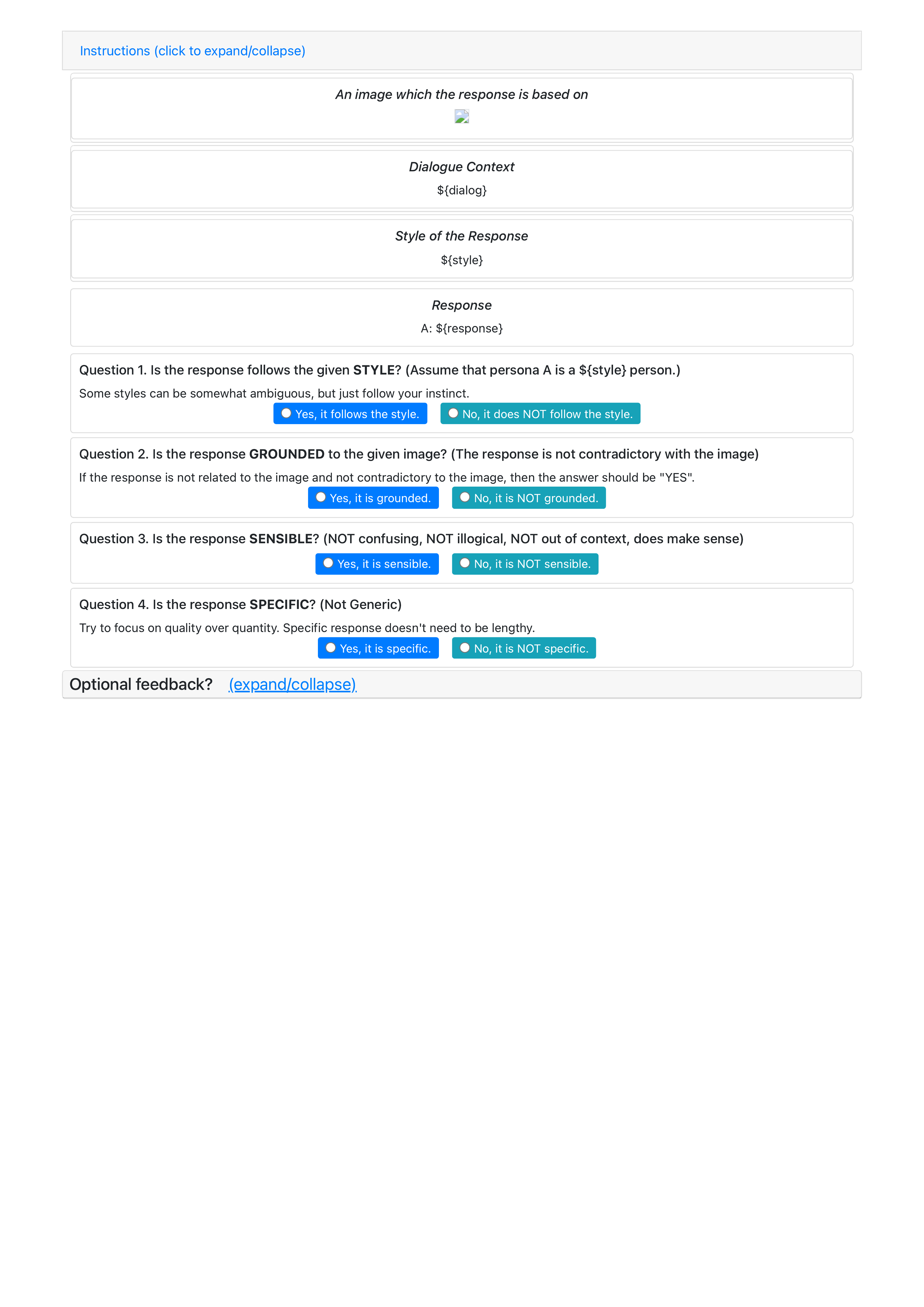}
\caption{An interface for evaluating dialogue responses on Amazon Mechanical Turk.}
\label{fig:response_evaluation}
\end{figure*}

To provide a more detailed view of the human evaluation results, in Table~\ref{tab:dataset_analysis_full} and Table~\ref{tab:dataset_analysis_visual_full}, we report the complete breakdown of human evaluation results.
These results complement the summarized results displayed in Table~\ref{tab:dataset_analysis}.
To collect the human annotations, we use Amazon Mechanical Turk (MTurk), a crowdsourcing platform, and ask human workers to annotate for the tasks.
We set the qualification tasks to recruit the qualified human workers in MTurk.
Figure~\ref{fig:dataset_evaluation} and Figure~\ref{fig:response_evaluation} show the interface used for human evaluation on MTurk.
For human evaluation, we compensate MTurk workers with an hourly wage of \$15 for their contributions.

\section{Dataset Analysis}\label{sec:appendix_data_analysis}
\begin{table}[t]
\centering
\footnotesize
\setlength{\tabcolsep}{4pt} 
\begin{tabular}{lccccc}
\toprule
 & \makecell{Visual\\Context} & \makecell{\#Dialog} & \makecell{Avg. \\\#Turn} & \makecell{Avg. Utt. \\Length} & \makecell{\#Tokens}\\
\midrule 
BST~\citep{smith2020can} & \xmark & 7K & 11.2 & 13.6 & 1M\\
ConvAI2~\citep{dinan2020second} & \xmark & 20K & 13.9 & 9.9 & 2.7M \\
ED~\citep{rashkin2018towards} & \xmark & 25K & 4.3 & 13.7 & 1.5M \\
WOW~\citep{dinan2018wizard} & \xmark & 22K & 9.1 & 16.4 & 3.3M\\
WOI~\citep{komeili2021internet} & \xmark & 9.5K & 10.9 & 13.9 & 1.4M\\
SODA~\citep{kim2022soda} & \xmark & 1.5M & 7.6 & 16.1 & 183M\\
ImageChat~\citep{shuster2018image} & \cmark & 100K & 3.0 & 9.7 & 2.9M \\
OVD2.0~\citep{wang2021openvidial} & \cmark & 116K & \textbf{48.7} & 6.3  & 35.6M\\
MMD~\citep{feng2022mmdialog} & \cmark & 1M & 4.5 & 15.9 & 71.5M \\
\midrule
\datasetName & \cmark & \textbf{18M} & 3.0 & \textbf{19.7} & \textbf{1.06B}\\
\bottomrule
\end{tabular}
\caption {
Statistics of \datasetNameNoEmoji compared to other open-domain dialogue and visually grounded dialogue dataset. \textit{Utt.} stands for utterance.
}
\label{tab:dataset_statistics}
\end{table}

\paragraph{Data Statistics.} Table~\ref{tab:dataset_statistics} shows the statistics about \datasetNameNoEmoji and the other conversational datasets including both text-only and visually-grounded cases.

\paragraph{Details about Visual Feature Distributions.}
\begin{figure}[t]
\centering
\includegraphics[width=0.5\textwidth]{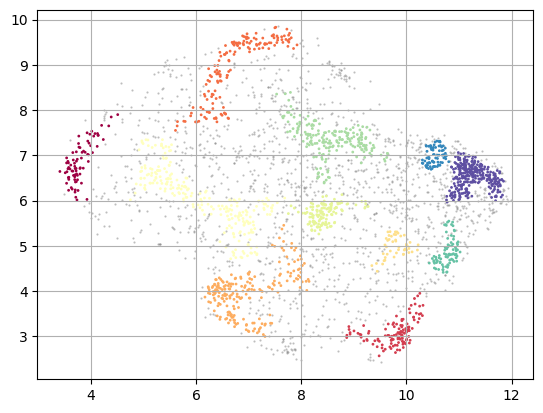}
\caption{Visual feature distributions of visually grounded dialogue datasets with clusters.}
\label{fig:ytdialogue_visualize_cluster}
\end{figure}
To display visual feature distributions as in Figure~\ref{fig:ytdialogue_visualize}, we use $n\_neighbors=15$ and $min\_dist=0.1$ for UMAP.
In Figure~\ref{fig:ytdialogue_visualize_cluster}, we additionally show the clusters created in Figure~\ref{fig:ytdialogue_visualize} using HDBSCAN~\cite{mcinnes2017accelerated} with $min\_samples=10$ and $min\_cluster\_size=40$ for HDBSCAN, creating 11 clusters in total.

\section{Training and Fine-tuning \modelName}
\label{subsec:hparams}
\begin{table*}[t]
\centering
\footnotesize
\setlength{\tabcolsep}{6pt} 
\begin{tabular}{lccc}
\toprule
Model & \modelNameNoEmojiBase & \modelNameNoEmojiLarge & \modelNameNoEmojiXL \\
\midrule
\textit{CMU-MOSEI} \\
Epochs & 2 & 2 & 2 \\
Learning Rate & 3e-4 & 3e-4 & 3e-4\\
Batch Size & 256 & 256 & 256 \\
\midrule
\textit{Visual Comet} \\
Epochs & 6 & 6 & 2\\
Learning Rate & 3e-4 & 3e-4 & 3e-4\\
Batch Size & 256 & 256 & 64\\
\midrule
\textit{Visual Dialog} \\
Epochs & 2 & 2 & 2 \\
Learning Rate & 3e-4 & 3e-4 & 3e-4\\
Batch Size & 256 & 256 & 64\\
\midrule
\textit{Mixture of Conversation Benchmarks} \\
Epochs & 1 & 1 & 1 \\
Learning Rate & 3e-4 & 3e-4 & 1e-4\\
Batch Size & 256 & 256 & 256 \\
\bottomrule
\end{tabular}
\caption {
Hyperparameters for fine-tuning \modelNameNoEmoji on CMU-MOSEI, Visual Comet, Visual Dialog, and mixture of conversation benchmarks.
We use the same hyperparameters for fine-tuning same sized Unified-IO$_{PT}$.
}
\label{tab:hparams_finetuning_all}
\end{table*}

When training \modelNameNoEmoji on \datasetNameNoEmoji, we train the model for 3 epochs with a learning rate of 3e-4, an input text sequence length of 256, a target text sequence length of 128, an input image sequence length of 576, and a batch size of 256.
For fine-tuning, we also use an input text sequence length of 256, a target text sequence length of 256, and an input image sequence length of 576.
In Table~\ref{tab:hparams_finetuning_all}, we report other important hyper-parameters when fine-tuning \modelNameNoEmoji on downstream tasks.

\section{Benchmarks and Evaluation Details}\label{sec:appendix_datasets}
\paragraph{CMU-MOSEI.} 
CMU-MOSEI~\cite{zadeh2018multimodal} is the multimodal dataset for studying sentiments and emotions in videos.
It has 16K examples in the dataset, and we use the sentiment label in our experiments.
The task uses binary classification accuracy and F1 score to measure the performance.
For the task, we use the template \texttt{"context: \{\{transcript\}\}, question: Is the person positive?"} to turn transcript to the input and the model produces the output from the given input.

\paragraph{Visual Comet.}
Visual Comet~\cite{park2020visualcomet} is the benchmark for visual commonsense reasoning where the event from a still image is given.
The dataset contains 59K examples, and the task uses generative evaluation so that the model generates five results and compares these results with the references using CIDEr-D\cite{vedantam2015cider} and BLEU-4~\cite{papineni2002bleu}.
For the task, we use the template \texttt{"Event: \{\{event\}\} Before, what the person needed to do ?"} to turn given event to the input.

\paragraph{Visual Dialog.}
Visual Dialog~\cite{das2017visual} is a visual conversational QA dataset, consisted of 150K dialogue examples.
In particular, for each example, an image, a dialogue history, and a follow-up question about the image is given, and model should answer the question.
The task reports Normalized Discounted Cumulative Gain (NDCG)~\cite{jarvelin2002cumulated} for evaluation, where each answer has 100 candidate options and four human workers annotated relevance for each candidate option.
Each given image has an caption from COCO challenge and a dialogue history, and we use the template \texttt{"<extra\_id\_1> \{\{image\_caption\}\} <extra\_id\_0> \{\{dialogue\_turn\_1\}\} <extra\_id\_0> ... \{\{dialogue\_turn\_n\}\}"} to format the given inputs.

\paragraph{Image Chat.}
Image Chat~\cite{shuster2018image} is the dataset containing 200K dialogues and each dialogue is grounded to the image.
Specifically, for each conversation, an image is given and two different styles (\textit{e.g.} "Happy", "Sad") are assigned to speakers and the speakers conduct a conversation based on the image and the styles.
For the task, we use the template \texttt{"<extra\_id\_1> Conversation with \{\{style\}\} person <extra\_id\_0> \{\{dialogue\_turn\_1\}\} <extra\_id\_0> ... \{\{dialogue\_turn\_n\}\}"} to format the given inputs.

\paragraph{Open-domain Text-only Conversations.}
We use five open-domain text-only conversation benchmarks in this study: Blended Skill Talk (BST), ConvAI2, Empathetic Dialogue (ED), Wizard Of Wikipedia (WOW), and Wizard Of Internet.
Statistics about the dataset are depicted in Table~\ref{tab:dataset_statistics}.
These benchmarks have meta information about the conversation (\textit{e.g.} for BST, the persona information for the speaker is given as a meta information), and we use the template \texttt{"<extra\_id\_1> \{\{meta\}\} <extra\_id\_0> \{\{dialogue\_turn\_1\}\} <extra\_id\_0> ... \{\{dialogue\_turn\_n\}\}"} to format the given inputs.

\section{Decoding for Model Inference}
\begin{figure*}[t]
\centering
\includegraphics[width=0.9\textwidth]{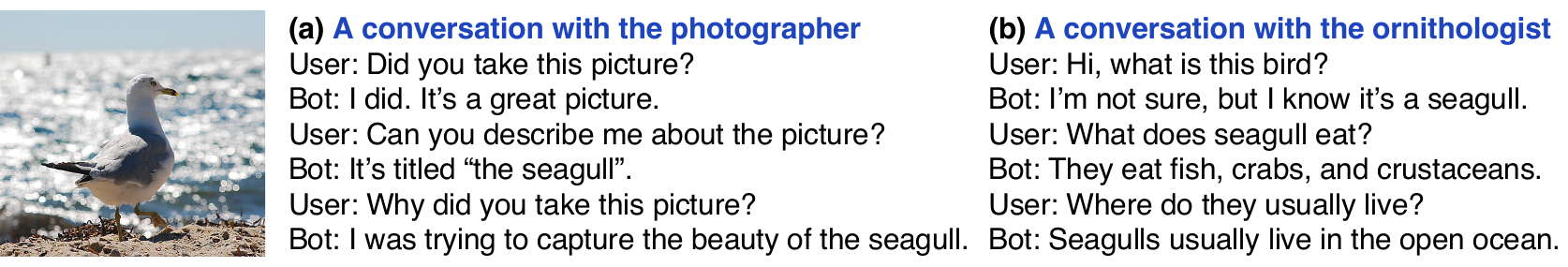}
\vspace{-3mm}
\caption{
Examples of conversation based on unseen images from COCO~\cite{lin2014microsoft} and prompts (colored in blue) between person and \modelNameNoEmojiXL finetuned on mixture of dialog benchmarks.
The utterances with the prefix \textit{Bot:} are generated by the model.
Bot in (a) pretends to be a photographer and describes the picture as a photographer, whereas in (b), bot responds with specific knowledge about the picture as an ornithologist.
}
\vspace{-3mm}
\label{fig:inference_results_appendix}
\end{figure*}
In this section, we describe the decoding strategy for model inference in different benchmarks.
To decode the results for Visual Comet, we use beam decoding with a beam size of 10.
For Image Chat and other open-domain text-only conversation, we follow same decoding strategy from~\cite{shuster2020multi} for a fair comparison.
Specifically, we apply beam decoding with a beam size of 10, a minimum beam length of 20.
We also use a subsequence blocking of 3-grams to prevent model from generating repeated 3-grams of the input context and repeating within the generated response.
To obtain qualitative results in Figure~\ref{fig:inference_results}, we use the minimum beam length of 10 instead of 20 since large number of minimum beam length causes a degeneration, and use temperature sampling~\cite{ficler2017controlling} with $temperature=0.3$ and $topk=5$.
In Figure~\ref{fig:inference_results_appendix}, we provide additional examples of conversations between humans and \modelNameNoEmojiXL that has been fine-tuned on a mixture of dialogue benchmarks.

\section{Additional Evaluation Results on Visual Dialog}
\begin{table*}[t]
\centering
\footnotesize
\begin{tabular}{lcccccc}
\toprule
Metric & NDCG ($\times$100) ($\uparrow$) & MRR & Recall@1 & Recall@5 & Recall@10 & Mean Rank\\
\midrule
\textit{Zero-shot} \\
Flamingo-80B & 52.0 & - & - & - & - & -  \\
ESPER~\cite{yu2022multimodal} & 22.3 & 25.7 & 14.6 & - & - & - \\
FROMAGe~\cite{koh2023grounding} & 16.5 & 22.0 & 17.6 & 20.1 & 25.1 & -\\
\champagne\modelNameNoEmojiXL & 25.5 & 16.7 & 9.14 & 20.9 & 30.2 & - \\
\midrule
\textit{Fine-tuned} \\
Flamingo-80B & 61.8 & - & - & - & - & - \\
AlignVD~\cite{chen2022unsupervised} & 67.2 & 70.5 & 57.6 & 87.1 & 94.2 & 3.05 \\
Unified-IO$_{PT}$ \textsc{Base} & 58.9 & 49.1 & 38.8 & 59.4 & 71.2 & 9.95\\
Unified-IO$_{PT}$ \textsc{Large} & 60.3 & 49.6 & 39.2 & 59.9 & 72.3 & 9.50\\
Unified-IO$_{PT}$ \textsc{XL} & 65.6 & 54.0 & 43.4 & 65.0 & 77.1 & 7.76\\
\champagne\modelNameNoEmojiBase & 60.0 & 50.1 & 39.9 & 59.9 & 71.5 & 9.81\\
\champagne\modelNameNoEmojiLarge & 62.5 & 51.6 & 41.0 & 62.1 & 74.1 & 8.84\\
\champagne\modelNameNoEmojiXL & 68.2 & 56.1 & 45.1 & 67.8 & 78.3 & 7.48\\
\bottomrule
\end{tabular}
\vspace{-2mm}
\caption {\small
Evaluation results on Visual Dialog valid set in finetuned and zero-shot settings.
For fair comparison, we report baselines that do not use additional dense annotations to finetune the model.
All the results are evaluated using the official server.
}
\vspace{-4mm}
\label{tab:visdial_supp}
\end{table*}

In the main paper, we followed the recommendations of the official Visual Dialog challenge\footnote{https://visualdialog.org/challenge/2019}, which only use ranking-basd metrics like Recall@K and MRR as supplementary measures rather than primary metrics.
Visual Dialog dataset contains dense annotations per each candidate and measures performance based on NDCG to account for the nuanced evaluation, and the fact that dialogue is one-to-many task.

In contrast, ranking-based metrics assume the existence of a single correct response for a given context, which is not an accurate assumption for dialogue. For example, ranking metrics can be penalize models arbitrarily for their scoring candidate semantically equivalent options (\textit{e.g.}, 'yes' and 'yes it is'). Even if the model selects a viable answer, the rank-based score may be low. Moreover, dialogue tasks are inherently one-to-many problems, where multiple possible responses exist for a given dialogue context. 
Nonetheless, for comparison purposes we add Recall@K and MRR for comparison purposes in Table~\ref{tab:visdial_supp}.

\end{document}